\documentclass[lettersize,journal]{IEEEtran}
\usepackage{amsmath,amsfonts}
\usepackage{array}
\usepackage[caption=false,font=normalsize,labelfont=sf,textfont=sf]{subfig}
\usepackage{textcomp}
\usepackage{stfloats}
\usepackage{url}
\usepackage{verbatim}
\usepackage{graphicx}
\usepackage{cite}
\newcommand{\ie}[1]{\textit{i.e.}}
\newcommand{\eg}[1]{\textit{e.g.}}

\usepackage[table]{xcolor}
\definecolor{coldgrey}{RGB}{128,128,105}
\usepackage{bbding}
\usepackage{pifont}

\usepackage[ruled,linesnumbered]{algorithm2e}
\usepackage{multicol}
\usepackage{multirow}
\usepackage{booktabs}
\usepackage{svg}

\begin{document}

\title{Rethinking Client Drift in Federated Learning: A Logit Perspective}

\author{Yunlu Yan, Chun-Mei Feng, Mang Ye, \IEEEmembership{Senior Member, IEEE}, Wangmeng Zuo, \IEEEmembership{Senior Member, IEEE}, Ping~Li, Rick Siow Mong Goh, Lei Zhu, 
		~C.~L.~Philip~Chen,~\IEEEmembership{Fellow,~IEEE,}

\thanks{Yunlu Yan is with The Hong Kong University of Science and Technology (Guangzhou), Nansha, Guangzhou, 511400, China.~(Email: yyan538@connect.hkust-gz.edu.cn)}
\thanks{Chun-Mei Feng and and Rick Siow Mong Goh, are with the
Institute of High Performance Computing, A*STAR, Singapore, 138632,    China.~(Email: strawberry.feng0304@gmail.com, gohsm@ihpc.a-star.edu.sg).}
\thanks{Mang Ye is with the Hubei Luojia Laboratory, National Engineering
Research Center for Multimedia Software, School of Computer Science,
Wuhan University, Wuhan, 430072, China. ~(Email: mangye16@gmail.com)}
\thanks{Wangmeng  Zuo is with the School of Computer Science and Technology, Harbin
Institute of Technology, Harbin, 130407, China. ~(Email: wmzuo@hit.edu.cn)}

\thanks{Lei Zhu is with The Hong Kong University of Science and Technology (Guangzhou), Nansha, Guangzhou, 511400, China and The Hong Kong University of Science and Technology, Hong Kong SAR, China.~(Email: leizhu@ust.hk)}
\thanks {Ping Li is with the Department of Computing and the School of Design, The Hong Kong
Polytechnic University, Hong Kong, China. (Email: p.li@polyu.edu.hk).}
			\thanks{C. L. Philip Chen is with the School of Computer Science and Engineering, South China University of Technology, Guangzhou, 510006, China, and also with the Pazhou Lab, Guangzhou, 510335, China (Email: philip.chen@ieee.org).}
\thanks{Lei Zhu (leizhu@ust.hk) is the corresponding author of this work.}}

\markboth{Journal of \LaTeX\ Class Files,~Vol.~14, No.~8, August~2021}%
{Shell \MakeLowercase{\textit{et al.}}: A Sample Article Using IEEEtran.cls for IEEE Journals}


\maketitle

\begin{abstract}
Federated Learning (FL) enables multiple clients to collaboratively learn in a distributed way, allowing for privacy protection. However, the real-world non-IID data will lead to client drift which degrades the performance of FL. 
Interestingly, we find that the difference in logits between the local and global models increases as the model is continuously updated, thus seriously deteriorating FL performance. This is mainly due to catastrophic forgetting caused by data heterogeneity between clients.
To alleviate this problem, we propose a new algorithm, named FedCSD, a \underline{C}lass prototype \underline{S}imilarity \underline{D}istillation in a federated framework to align the local and global models.
FedCSD does not simply transfer global knowledge to local clients, as an undertrained global model cannot provide reliable knowledge, i.e., class similarity information, and its wrong soft labels will mislead the optimization of local models. Concretely, FedCSD introduces a class prototype similarity distillation to align the local logits with the refined global logits that are weighted by the similarity between local logits and the global prototype. To enhance the quality of global logits,  FedCSD adopts an adaptive mask to filter out the terrible soft labels of the global models, thereby preventing them to mislead local optimization. Extensive experiments demonstrate the superiority of our method over the state-of-the-art federated learning approaches in various heterogeneous settings. The source code will be released.
\end{abstract}


\section{Introduction}

Federated Learning (FL)~\cite{yang2019survey,li2020survey} is an emerging distributed learning paradigm that has garnered substantial interest, particularly in privacy-sensitive domains like healthcare~\cite{feng2021specificity,liu2021feddg,karargyris2023federated, wicaksana2022fedmix,wang2023feddp,yan2023cross}, where it enables multiple clients to collaboratively train machine learning models while maintaining privacy and avoiding data exposure. However, when FL is applied to a multitude of discrete clients, the datasets associated with each client in real-world scenarios inevitably stem from distinct underlying distributions, resulting in non-IID data. This non-IID data phenomenon can lead to what is known as \textbf{\textit{client drift}}~\cite{karimireddy2020scaffold}, which subsequently undermines the performance of FL~\cite{li2019convergence,wang2020tackling,zhao2018federated}.

\begin{figure}[t]
\centering
  \includegraphics[width=0.5\textwidth]{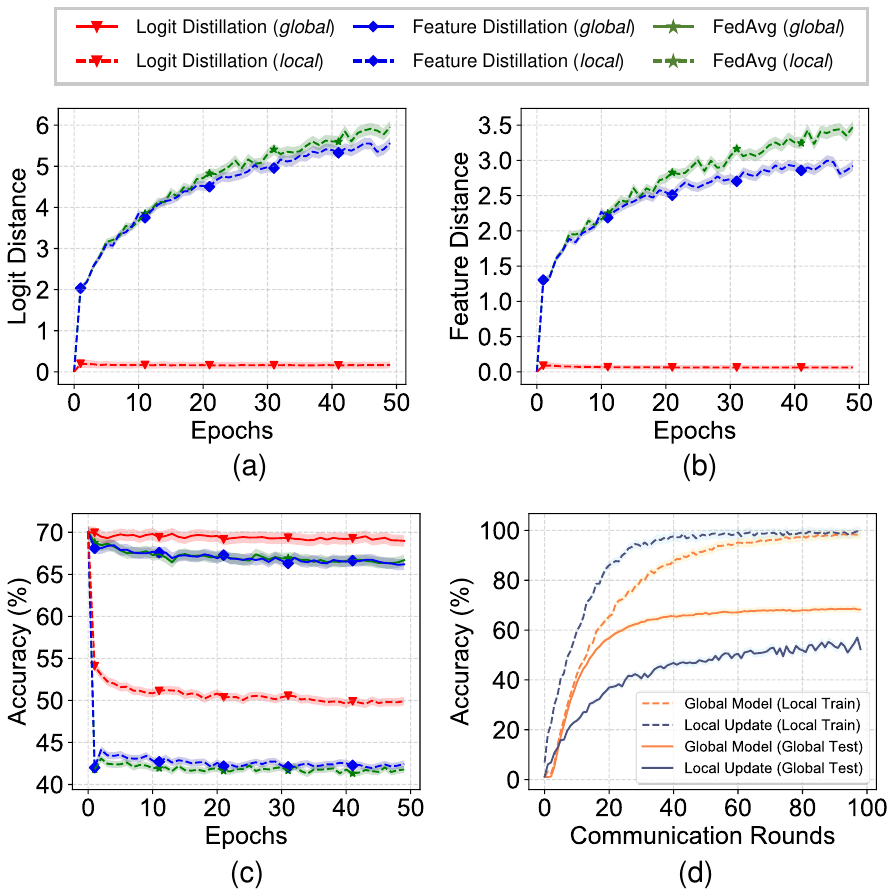} \caption{\textbf{Evolution illustration} of global and local models on \textbf{CIFAR-100}~\cite{krizhevsky2009learning} versus \textbf{(a)} \textit{logit distance}, \textbf{(b)} \textit{feature distance}, \textbf{(c)} their \textit{accuracy} on the exploratory experiment, and \textbf{(d)} their \textit{accuracy}  on both the global test set and local training set.}
  \label{fig:lcurve} 
\end{figure}

Non-IID data is commonly encountered in various real-world scenarios, prompting many researchers to address this challenge to enhance the viability of federated learning. Existing solutions can be categorized into two types: \textbf{\textit{client-specific learning}}~\cite{li2021fedbn,wang2023feddp,li2021ditto,tan2022towards} and \textbf{\textit{client-unified learning}}~\cite{li2020federated,karimireddy2020scaffold,wang2020tackling,li2021model,zhang2022federated}. The former solves the problem through the personalized design for each client to learn the specific model representation for each client, \textit{e.g.}, FedBN~\cite{li2021fedbn} and FedPer~\cite{arivazhagan2019federated}. In contrast, the latter approaches seek to develop a unified model representation that applies across all clients, which is explored in this work. In pursuit of this objective, researchers employ strategies such as regularization~\cite{li2020federated,acarfederated,gao2022feddc}, contrastive learning~\cite{li2021model}, and data augmentation~\cite{zhou2023fedfa,yan2023simple} to minimize the divergence among individual local models during the training process. While client-unified learning has been investigated in FL, it's intriguing to note that there has been limited attention directed towards the logits, \textit{i.e.}, the output of the classifier, which effectively captures the decision-making process of the classifier. Previous research~\cite{luo2021no} indicates that the heterogeneity among different local models primarily resides in the classifier, as evidenced by experimental exploration. Consequently, we contend that the discrepancy in logits is likely to be more pronounced than that in latent features. Therefore, addressing client drift directly and effectively through logits appears to be a promising approach.

In this work, a fortunate discovery was made, revealing the following key insights: \ding{182} \textbf{Logit Shift Due to Local Updates}: We observed that local updates lead to differences in logits between the local and global models, a phenomenon we refer to as \textbf{\textit{logit shift}} (illustrated in Fig.~\ref{fig:lcurve} (a)). \ding{183} \textbf{Impact of Logit Shift on FL Performance}: A noteworthy finding was that the logit shift is closely linked to the performance deterioration observed in FL, as demonstrated in Fig.~\ref{fig:lcurve} (c). \ding{184} \textbf{Mitigation of Feature Shift through Logit Alignment}: Interestingly, we found that aligning the local and global logits can effectively alleviate the feature shift phenomenon~\cite{zhou2023fedfa}, as depicted in Fig.~\ref{fig:lcurve} (b). These discoveries collectively offer valuable insights into the dynamics of logit behaviour and its implications for FL performance, which have sparked our motivation to approach the non-IID problem from a fresh and innovative perspective, \ie, logit shift. 

Considering that each local model is initialized from the parameters of the global model, we argue that the phenomenon of logit shift can be attributed to \textbf{\textit{catastrophic forgetting}}~\cite{kemker2018measuring,serra2018overcoming,boschini2022class}. This occurs when the local model trains on its private dataset which is from a biased distribution. Over the course of continuous training, the local model gradually relinquishes its initial grasp on global knowledge and becomes predisposed to the specifics of its local data. Consequently, this bias causes the local logits to deviate from the global logits. Drawing from these insights, a straightforward solution emerges: maintaining consistency between local and global model logits through knowledge distillation. This technique, commonly employed to mitigate catastrophic forgetting in continual learning scenarios~\cite{phan2022class, gao2022r}, offers a pragmatic strategy for mitigating the logit shift phenomenon.

While several studies~\cite{lee2021preservation, yao2021local} have incorporated knowledge distillation into Federated Learning (FL) by employing the global model as the teacher to regulate local optimization, we have identified a crucial aspect that these approaches overlook: \textbf{\textit{the global model is inadequately trained to serve as an effective teacher for a local model during local training}}, despite containing global knowledge. As illustrated in Fig.~\ref{fig:lcurve} (d), due to fine-tuning on the local dataset, the local model exhibits enhanced performance on the local training set. Meanwhile, the global model undergoes continual updates, distinguishing it from the conventional distillation process, where a well-trained teacher imparts its knowledge to a student. Consequently, direct alignment of local and global logits raises certain key challenges. \textit{\textbf{First}}, when applying knowledge distillation at the logits level, it's commonly believed that the teacher's logits serve as soft targets, transferring "dark knowledge" that includes privileged information regarding the relationships between different categories~\cite{hinton2015distilling}. This transfer is effective only when a stronger teacher imparts knowledge to a weaker student, as a poorly-trained teacher cannot reliably convey accurate similarity information among categories~\cite{yuan2020revisiting}. \textit{\textbf{Second}}, owing to the global model's lower accuracy, it generates a plethora of incorrect soft labels that misguide the optimization of local models, particularly in the early stages.

 Based on our analyses, we propose a novel FL framework, \ie, FedCSD, to tackle the non-IID data challenge from the vantage point of logit shift. FedCSD introduces a novel method termed class prototype similarity distillation, which aligns local logits with global logits, factoring in the similarity between local logits and the global prototype. Furthermore, we incorporate an adaptive mask mechanism to sieve out insignificant knowledge from global logits. By amalgamating these foundational elements, our approach adeptly resolves the non-IID problem in FL. In a nutshell, our contributions are summarized as follows:
 
\begin{itemize}
	\setlength{\itemsep}{4pt}
	\setlength{\parsep}{-2pt}
	\setlength{\parskip}{-0pt}
	\setlength{\leftmargin}{-15pt}
\item We provide a new perspective, \ie, the logit shift between local and global models, to help us understand the client drift under non-IID data, which is beneficial to handle this fundamental challenge. This also explains the underlying mechanism of our approach.
\item We propose FedCSD, a novel framework to address the client drift in FL. This framework employs a prototype-based class similarity distillation technique to align local and global logits, effectively curbing the occurrence of catastrophic forgetting within local models. Consequently, FedCSD serves as a potent strategy to alleviate the impact of client drift.

\item Extensive experiments on three typical FL datasets demonstrate the effectiveness of our method under various data heterogeneous settings, \eg, it outperforms various state-of-the-art FL approaches.
\end{itemize}

\section{Related work}

\noindent{\textbf{Federated Learning with Non-IID Data.}} The classical FL algorithm, Fedavg~\cite{mcmahan2017communication}, achieved a balance of computing and communication, which shows good performance in some applications~\cite{sun2022decentralized,dinh2021communication}. However, the accuracy of FedAvg reduces significantly when local data is non-IID~\cite{zhao2018federated,karimireddy2020scaffold,li2020federated}, which has been a fundamental challenge. To address this challenge, a variety of regularization methods~\cite{gao2022feddc,acar2020federated,li2020federated,karargyris2023federated,mendieta2022local} are used to enforce local optimization. For example, FedProx ~\cite{li2020federated} computed the $l_2$-norm distance between the weight of local and global models as a proximity term added to the local objective. Similarly,
FedDyn~\cite{acar2020federated} adopted a dynamic regularization into the local object based on exact minimization which seeks to keep the local-global optima consistent. Despite their efforts, the performance of FedAvg is not fully understood.
SCAFFOLD~\cite{karimireddy2020scaffold} provided a more delicate analysis of FedAvg for non-IID data and proves that client drift is the root of performance degradation. To solve the client drift problem, it introduced control variate to correct local updates. Besides, MOON~\cite{li2021model} proposed model-contrastive learning to correct the local training by utilizing the similarity between model representations among local and global. Some studies try to improve FedAvg in different ways. For example, bayesian non-parametric methods~\cite{wang2020federated}, momentum updating~\cite{hsu2019measuring}, normalize~\cite{wang2020tackling} are used to improve Fedavg on the phase of modal aggregation. However, the above methods ignore the key point of the potential influence of performance drop, \ie, logit, since the previous study has confirmed that the model drifts mainly focuses on the classifier layer~\cite{luo2021no}.
Instead, our work provides a novel perspective to address the client drift and achieve competitive results.

\noindent{\textbf{Knowledge Distillation.}}
Knowledge distillation~\cite{hinton2015distilling,gou2021knowledge} is a knowledge extraction and transfer paradigm by the teacher-student mechanism that attempts to transfer the knowledge from the teacher model into the student model.  Specifically, the logits of the teacher model as the soft labels that supervise the student model to train on a proxy dataset. Moreover, it aims to minimize the teacher-student logit discrepancy that can be measured by Kullback-Leibler divergence. 

Knowledge distillation has also been successfully applied in FL~\cite{sui2020feded,huang2022learn,chen2023metafed}. For example, some studies try to propose a communication-efficient FL framework based on knowledge distillation ~\cite{jeong2018communication,sattler2021cfd,wu2021fedkd}. Knowledge distillation has also been used to address the non-IID data. For instance, Seo \textit{et al.}~\cite{seo2020federated} assigned a client as a student which receives ensemble logits of the rest clients. FedDF~\cite{lin2020ensemble} used ensemble distillation to replace parameter averaging of FedAvg which needs extra training and a proxy dataset on the server.
FedGen~\cite{zhu2021data} utilized an additional generator to aggregate the local information which increases the additional training cost and privacy risk. Similar to us, Fed-NTD~\cite{lee2021preservation} distilled the knowledge of the not-true class between the local and global models. FedGKD~\cite{yao2021local} utilized several previous global models as an ensemble teacher to teach the local model. However, the performance of their method is limited by the poorly-trained global model, which can not provide reliable class similarity information and soft labels. To achieve effective distillation, we utilize two key modules, \ie, class prototype similarity distillation and adaptive mask, to improve the performance of the global model.

\begin{figure*}[t]
\centering
  \includegraphics[width=1.00\textwidth]{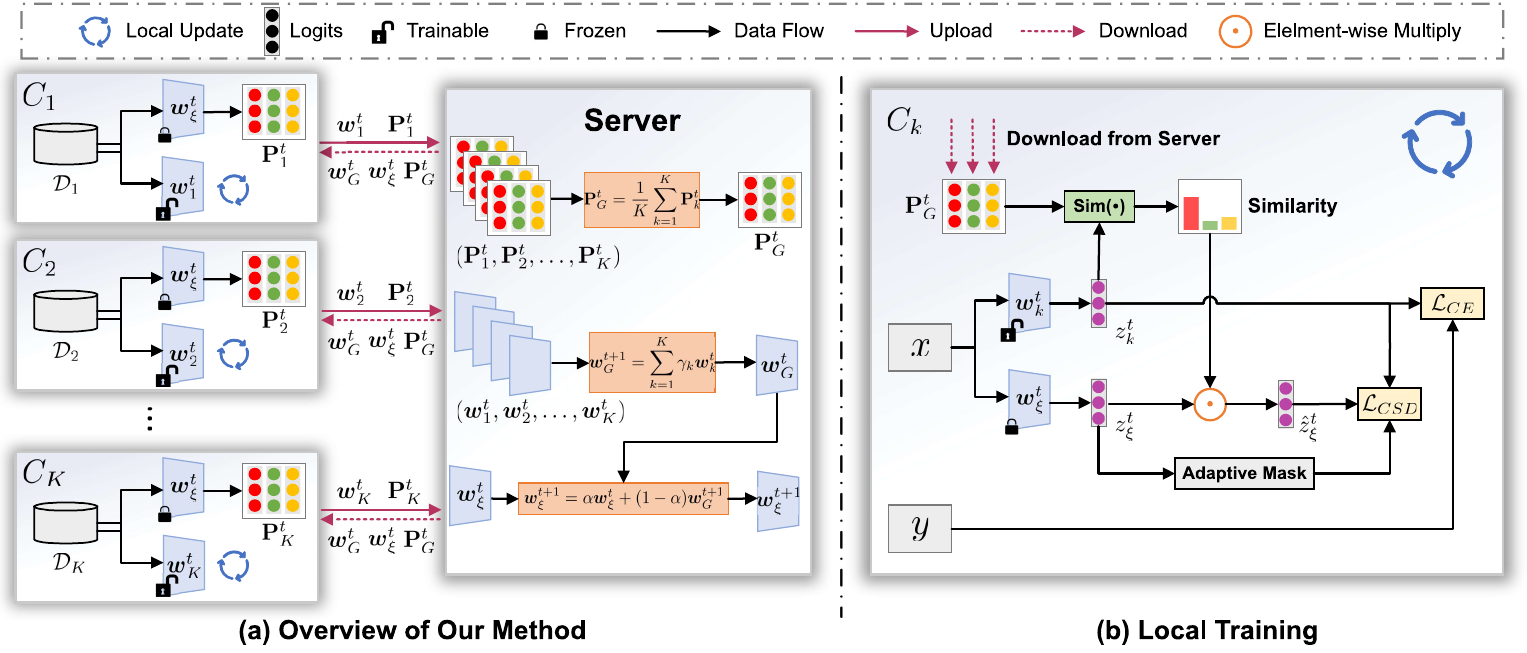}
  \caption{\textbf{Illustration of FedCSD}. (a) Before the local training, each client receives the global $\boldsymbol{w}_G^{t}$ and teacher $\boldsymbol{w}_\xi^{t}$ models to initialize the local model and compute the class prototype, and then the server receives the class prototype from each client to obtain the global class prototype $\mathbf{P}_G^t$. (b)   An adaptive mask and the global prototype
   are used to compute the distillation loss $\mathcal{L}_{CSD}$. The local objective is the weighted sum of the distillation loss $\mathcal{L}_{CSD}$ and cross-entropy loss $\mathcal{L}_{CE}$. }
  \label{fig:method_overview}
\end{figure*}

 \section{Preliminary}

 \subsection{Problem statement}
Assume a standard federation that there are $K$ clients $\{C_{1}, \ldots, C_{K}\}$ and a central server. Each client $C_k$ has $n_{k}$ training samples $\{x_i,y_i\}_{i=1}^{n_k}$, where image $x_i \in \mathcal{X}$ and corresponding label $y_i \in \mathcal{Y}$ are from a joint distribution, \ie, $(x_i,y_i) \sim \mathcal{D}_{k}(\mathcal{X}, \mathcal{Y})$. For non-IID data in FL, the distribution of each client $\mathcal{D}_{k}$ is different. The standard FL aims to learn a global optima model by minimizing the empirical loss of each client without privacy disclosure. The global objective function can be described as  ~\cite{mcmahan2017communication}:
\begin{equation}
\min \mathcal{L} = \sum_{k=1}^{K} \gamma_{k} \mathcal{L}_{k}
\label{eq:1},
\end{equation}
where $\gamma_{k}= \frac{n_{k}}{\sum_{i=1}^{K}n_{i}}$, $\mathcal{L}_{k}$ is the local objective function of $C_{k}$. Here, every client will learn a deep neural network $f$: $\mathcal{X} \rightarrow \mathcal{Y}$ via cross-entropy loss:
\begin{equation}
\mathcal{L}_{CE} = - \mathbb{E}_{(x_i,y_i)\sim \mathcal{D}_{k}}  \sum_{c\in\mathcal{Y}}^{}y_{i,c} \text{log}(\frac{\text{exp}(z_{i,c})}{\sum_{j\in\mathcal{Y}}^{}\text{exp}(z_{i,j})}) 
\label{eq:2}.
\end{equation}
Where logits $z_i =f(\boldsymbol{w}_{k};x_i)$ and $\boldsymbol{w}_{k}$ is the parameters of the local model. To improve communication efficiency, the leading algorithm FedAvg~\cite{mcmahan2017communication} conducts $E$ local epochs and then averages the local model parameters to update the global model parameters at each communication round:
\begin{equation}
\boldsymbol{w}_{G}^{t+1} =  \sum_{k=1}^{K} \gamma_{k} \boldsymbol{w}_{k}^{t}
\label{eq:3}.
\end{equation}
With the $T$ rounds training, we can get an optimal global model $\boldsymbol{w}_{*}$ which has the best global performance.

\subsection{Motivation}\label{sec:moti}

\noindent\textbf{Exploratory Experiment.} We conduct an intriguing experiment for FedAvg with the ResNet-50~\cite{he2016deep} network on the CIFAR-100~\cite{krizhevsky2009learning} dataset. In this experiment, we train a round with 50 local epochs involving 10 non-IID clients. Initially, each client receives a proficiently trained global model from the server, which serves as the basis for initializing the local model. After a single epoch of training, we assess both the logit distance and feature distance between the local update and the initial global model. The logit distance is quantified using the Kullback-Leibler divergence, while the feature distance is measured using the $l_2$-norm. Additionally, by averaging the parameters of local updates, we generate a global update and evaluate its performance on the test set. The outcome of the experiment is depicted in Fig.~\ref{fig:lcurve} (a, b, c), where \textbf{\textit{global}} signifies the result of the global update and \textbf{\textit{local}} represents the average of the results obtained from 10 local updates.

\noindent\textbf{Detail of Logit Distillation and Feature Distillation.}
The \textit{logit distillation}~\cite{hinton2015distilling} and \textit{feature distillation}~\cite{2014Fit} are two methods that distill the logit and feature of the global model to the student model, respectively. The loss of \textit{logit distillation} can be described as:
\begin{equation}
\begin{split}
    \mathcal{L}_{\mathrm{LD}} = - \mathbb{E}_{(x_i,y_i)\sim \mathcal{D}_{k}} \tau^2 \sum_{c\in \mathcal{Y}}^{}q_{G,c} \log (q_{k,c})
   , \quad \text{where}  \\ q_{G,c} \!= \!\frac{\text{exp}(z_{G,c}/\tau)}{\sum_{i\in\mathcal{Y}}^{}\!\text{exp}(z_{G,i}/\tau)} \! \ \
q_{k,c} \!= \!\frac{\text{exp}(z_{k,c}/\tau)}{\sum_{i\in\mathcal{Y}}^{}\!\text{exp}(z_{k,i}/\tau)}, 
\end{split} 
    \label{eq:lcsd}
\end{equation}
where, $z_{G}$ and $z_{k}$ are the logits of the global and local models, $\tau$ is the temperature hyper-parameter, and set to $10$ by default.

Besides, \textit{feature distillation} adopts the MSE loss to distill the feature of the global model:
\begin{equation}
    \mathcal{L}_{\mathrm{FD}} = (h_k - h_G)^2, 
\end{equation}
where $h_G$ and $h_k$ are the latent features of the global and local models, respectively. For ResNet-50, they are $1\times D$ vector, and D is $2048$. We fine-tune the $\mu$ from \{0.001, 0.01, 0.1, 1\}, and the optimal $\mu$ of \textit{logit distillation} and \textit{feature distillation} is empirically set to $0.001$ and $0.01$, respectively.

\noindent\textbf{Experiment Observation.} The results demonstrate a noteworthy trend: an increase in the number of local epochs contributes to a higher discrepancy in logits between the local update and the global model. This divergence arises due to the local model gradually losing its prior knowledge, which, in turn, adversely affects the accuracy of both local and global updates. Additionally, a rise in the number of local epochs also amplifies the feature distance, aligning with the observations from prior research~\cite{pengfederated} that non-IID data introduces differences in features. Intriguingly, we delve deeper by applying feature distillation and logit distillation techniques to align the features and logits of local and global models. Remarkably, aligning logits not only enhances the accuracy of both local and global updates but also indirectly maintains feature consistency. Importantly, logit distillation outperforms feature distillation in reducing the occurrence of forgetting. Consequently, we arrive at a significant conclusion: \textbf{\textit{Consistency between local and global logits is beneficial for both model aggregation and local optimization}}.

\subsection{Federated Learning and Continual Learning}
We analyze the relationships between FL and continual learning to further explain the intriguing results of the exploratory experiment. 
Considering a continual learning task, and given a well-trained model $\boldsymbol{w}_{old}$ on the dataset $\mathcal{D}_{old}$, the goal is to continually train the model on a new dataset $\mathcal{D}_{new}$ as preserving the learned knowledge. 
And we can suppose that the parameters of the model after the training on $\mathcal{D}_{new}$ is $\boldsymbol{w}_{old}+\sigma$, where $\sigma$ is the offset before and after the update. 
Due to the catastrophic forgetting, the performance of the new model on $\mathcal{D}_{old}$ will greatly drop, which reveals the difference between $f(\boldsymbol{w}_{old})$ and $f(\boldsymbol{w}_{old}+\sigma)$, \ie, logit shift. 
Analogous to continual learning, we can donate the local model parameters as $\boldsymbol{w}_G+\hat{\sigma}$ after the local training, where $\hat{\sigma}$ is the local update.  For non-IID data, the distribution $\mathcal{D}_{k}$ is different from the global distribution $\mathcal{D}$, which will cause a catastrophic forgetting problem, and the catastrophic forgetting contributes to the logit shift between local and global models.

\section{FedCSD}

In this section, we propose a Federated Class Prototype Similarity Distillation (FedCSD) framework which contains two key components, \ie, class prototype similarity distillation and adaptive mask. Our focus is mainly on the local training phase and with light modification on the global aggregating phase. An overview of our method is illustrated in Fig. \ref{fig:method_overview} and Alg.~\ref{alg:fedCSD}. In the following, we will introduce the detail of our method.

\subsection{Class Prototype Similarity  Distillation} 

As previously mentioned, in terms of local training, the global model is a weak teacher for the local model though it has learned global knowledge from different clients. Due to the weaker performance of the global model on the local dataset, it can not provide reliable similarity information among the different classes for the local model. Therefore, to strengthen the class similarity information of the soft labels, \ie, global logits, we introduce a class prototype similarity weight to refine the soft labels.

\noindent\textbf{Class Prototype Generating:} As shown in Fig. \ref{fig:method_overview}, before the $t$-th round training at client $C_k$, it will receive the parameters $\boldsymbol{w}_{\xi}^{t}$ of the teacher. To begin with, the teacher calculates the logit $z_{\xi, i}^{t} \in \mathbb{R}^{1\times|\mathcal{Y}|}$ of each instance $(x_i, y_i) \in \mathcal{D}_k$.
Then, the teacher will obtain the prototype vector $\mathbf{P}_{k,c}^t \in \mathbb{R}^{1\times|\mathcal{Y}|}$ of class $c \in \mathcal{Y}$ by computing the in-dataset average on the logits $\{z_{\xi, i}^t\}^{n_k}_{i=1}$ as:
\begin{equation}
\mathbf{P}_{k,c}^t=\frac{\sum_{i=1}^{n_{k}} z_{\xi, i}^t \mathbb{I}\left[y_i=c\right]}{\left|\left\{i: y_i=c\right\}\right|}, \ \text{where} \ z_{\xi, i}^t = f(\boldsymbol{w}_{\xi}^{t};x_i), 
\label{eq:lp}
\end{equation}
where $\mathbb{I}[\cdot]$ is the indicator function, $\mathbb{I}[y_i=c]$ is $1$ if the label $y_i$ is equal to $c$ and $0$ otherwise. For non-IID data in FL, the prototype matrix $\mathbf{P}_k^t = [\mathbf{P}_{k, 1}^t, \mathbf{P}_{k, 2}^t, ..., \mathbf{P}_{k, |\mathcal{Y}|}^t]$,  $\mathbf{P}_k^t \in \mathbb{R}^{|\mathcal{Y}|\times|\mathcal{Y}|}$  may not contain all categories, especially for the label skew~\cite{li2022federated}, a typical non-IID situation. Moreover, the class prototype $\mathbf{P}_k^t$ only learns the information of $C_k$ and lacks cross-client consistency. Therefore, we send the local class prototype to the server, and then the server aggregate all local class prototypes to obtain the global class prototype $\mathbf{P}_G^t$, which can be described as:
\begin{equation}
\mathbf{P}_G^t=\frac{1}{K}\sum_{k=1}^{K}\mathbf{P}_k^t.
\end{equation}
\noindent\textbf{Privacy Preserving:} In particular, the prototype does not contain any information related to privacy~\cite{tan2022fedproto,tan2022federated,Huang_2023_CVPR}, and also can not be reversed to the individual image because it is statistical information of the whole dataset.

\begin{algorithm}[!t]
\caption{FedCSD}
\label{alg:fedCSD}
\KwIn{$K$ local datasets: $\{\mathcal{D}_{1}, \mathcal{D}_{2},  \ldots, \mathcal{D}_{K}\}$, communication rounds $T$, local epochs $E$, temperature $\tau$, learning rate $\eta$, loss weight $\mu$,
momentum hyper-parameter $\alpha$}
\KwOut{$\boldsymbol{w}^{T}_{G}$}
initialize $\boldsymbol{w}^{0}_{G}$, $\boldsymbol{w}^{0}_{\xi}$ \\
\For{round $t=0,1,...,T-1$}
{
    \For{client $k=1,2,...,K$ \textbf{parallelly}}{
        $\mathbf{P}_k^t$ $\leftarrow$ $\{\frac{\sum_{i=1}^{n_{k}} z_{\xi, i}^t \mathbb{I}\left[y_i=c\right]}{\left|\left\{i: y_i=c\right\}\right|}\}_{c\in\mathcal{Y}}$
    }

    $\mathbf{P}_G^t=\frac{1}{K}\sum_{k=1}^{K}\mathbf{P}_k^t$
    
    \For{client $k=1,2,...,K$ \textbf{parallelly}}{
    $\boldsymbol{w}_{k}^{t}$ $\leftarrow$ Local Training ($k$, $\mathbf{P}_G^t$, $\boldsymbol{w}_{\xi}^{t}$, $\boldsymbol{w}_{G}^{t}$)
}
$\boldsymbol{w}_{G}^{t+1} =  \sum_{k=1}^{K} \gamma_{k} \boldsymbol{w}_{k}^{t}$ \\

$\boldsymbol{w}_{\xi}^{t+1} = \alpha\boldsymbol{w}_{\xi}^{t} +(1-\alpha)\boldsymbol{w}_{G}^{t+1}$ \\

} 
return $\boldsymbol{w}_{G}^{T}$

\textbf{Local Training} ($k$, $\mathbf{P}_G^t$, $\boldsymbol{w}_{\xi}^{t}$, $\boldsymbol{w}_{G}^{t}$):\\
    $\boldsymbol{w}_{k}^{t}$ $\leftarrow$ $\boldsymbol{w}_{G}^{t}$ \\ 
\For{epoch $e=1,2,...,E$}{
        
        \For{batch $b=(x,y)$ $\sim$ $\mathcal{D}_{i}$}{
        \resizebox{0.8\linewidth}{!}{$\mathcal{L}^k
        = \mathcal{L}_{CE}^k(\boldsymbol{w}_{k}^{t};x;y)+\mu
\mathcal{L}_{CSD}^k(\mathbf{P}_G^t;\boldsymbol{w}_{\xi}^{t};\boldsymbol{w}_{k}^{t};x;y)$} \\
$\boldsymbol{w}_{k}^{t}$ $\leftarrow$ $\boldsymbol{w}_{k}^{t} - \eta \nabla \mathcal{L}^k$
       }
    }
return $\boldsymbol{w}_{k}^{t}$
\end{algorithm}

\noindent\textbf{Similarity Estimation:} After obtaining the global class prototype, each client can download $\mathbf{P}_G^t$ from the server to calculate the cosine similarity $\delta \in \mathbb{R}^{1\times|\mathcal{Y}|}$ between the local logits and global class prototype during the local training:
\begin{equation}
\delta =\frac{z_k^t \cdot \mathbf{P}_G^t}{\left\|z_k^t\right\| \times\left\|\mathbf{P}_G^t\right\|}, \ \text{where} \ z_k^t = f(\boldsymbol{w}_{k}^t; x_i).
\end{equation}
We further normalized the cosine similarity $\delta = \{\delta_c\}_{c\in\mathcal{Y}}$ to get the similarity score $\hat{\delta}$, which is defined as:
\begin{equation}
\hat{\delta}_c=\frac{\text{exp}(\delta_c)}{\sum_{i\in\mathcal{Y}}^{}\text{exp}(\delta_i)}, \quad  \text{where} \ c \in \mathcal{Y}.
\end{equation}
With the above normalization, the range of $\hat{\delta}$ is changed to $[0, 1]$ and we utilize it to refine the logits of the teacher to enhance the class similarity information:
\begin{equation}
\hat{z}_\xi^t = \hat{\delta} z_\xi^t, \ \text{where} \ z_\xi^t = f(\boldsymbol{w}_{\xi}^t; x_i).
\end{equation}
In the following, we introduce the knowledge distillation to align the weighted teacher logits and the local logits as preserving the global knowledge of the local model. The class prototype similarity distillation loss can be written as:
\begin{equation}
\begin{split}
    \mathcal{L}_{\mathrm{CSD}} = - \mathbb{E}_{(x_i,y_i)\sim \mathcal{D}_{k}} \tau^2 \sum_{c\in \mathcal{Y}}^{}q_{\xi,c}^t \log (q_{k,c}^t)
   , \quad \text{where}  \\ q_{\xi,c}^t \!= \!\frac{\text{exp}(\hat{z}_{\xi,c}^t/\tau)}{\sum_{i\in\mathcal{Y}}^{}\!\text{exp}(\hat{z}_{\xi,i}^t/\tau)} \! \ \
q_{k,c}^t \!= \!\frac{\text{exp}(z_{k,c}^t/\tau)}{\sum_{i\in\mathcal{Y}}^{}\!\text{exp}(z_{k,i}^t/\tau)}, 
\end{split}
    \label{eq:lcsd}
\end{equation}
where $\tau$ is the temperature hyper-parameter.

\noindent\textbf{Teacher Update:} It is worth noting that the teacher is frozen during the local training. Besides, the previous works directly used the global model of the last round as the teacher and the quality of the teacher will be impacted by the unsteady training process. Hence, to provide a more stable teacher, we utilize a common model smoothing technology, \ie, Temporal Moving Average~\cite{tarvainen2017mean} (TMA), to update the teacher. Specifically, it utilizes the global models for different time periods to obtain the teacher by momentum update, which can be defined as:
\begin{equation}
\boldsymbol{w}_\xi^{t+1} = \alpha\boldsymbol{w}_\xi^{t} + (1-\alpha)\boldsymbol{w}^{t+1}_{G}, \quad \boldsymbol{w}_\xi^0 = \boldsymbol{w}_G^0,
\label{eq:tma}
\end{equation}
where $\alpha$ is the momentum hyper-parameter.

\begin{table*}[h]
\centering
\caption{\textbf{ The test accuracy (\%) of all approaches with different $\beta$ on CIFAR-100~\cite{krizhevsky2009learning} and FEMNIST~\cite{caldas2018leaf}}. ${\color{green}\uparrow}$ and ${\color{red}\downarrow}$ show the rise and fall compared with FedAvg. We mark the best results in bold.}
 \small
\setlength\tabcolsep{8pt}
\resizebox{0.95\textwidth}{!}{
\renewcommand\arraystretch{1.5}{
\begin{tabular}{lllllll}
\toprule
\multirow{2}{*}{\textbf{Method}} & \multicolumn{3}{c}{\textbf{CIFAR-100}~\cite{krizhevsky2009learning}}
& \multicolumn{3}{c}{\textbf{FEMNIST}~\cite{caldas2018leaf}}\\
\cmidrule(lr){2-4}
\cmidrule(lr){5-7}
& $\beta=0.01$ & $\beta=0.5$ & $\beta=5$ & $\beta=0.01$ & $\beta=0.05$ & $\beta=0.5$\\
\midrule
FedAvg~\cite{mcmahan2017communication}       & $58.50_{\color{coldgrey}(base)}$  & $66.67_{\color{coldgrey}(base)}$  & $68.83_{\color{coldgrey}(base)}$ & $86.36_{\color{coldgrey}(base)}$ & $97.31_{\color{coldgrey}(base)}$ & $99.07_{\color{coldgrey}(base)}$ \\
\midrule
FedProx~\cite{li2020federated}    & $59.37_{\color{coldgrey}(0.87)}$ ${\color{green}\uparrow}$  & $68.64_{\color{coldgrey}(1.97)}$ ${\color{green}\uparrow}$  & $69.64_{\color{coldgrey}(0.81)}$ ${\color{green}\uparrow}$ & $76.40_{\color{coldgrey}(9.96)}$ ${\color{red}\downarrow}$ & $97.53_{\color{coldgrey}(0.22)}$ ${\color{green}\uparrow}$ & $99.24_{\color{coldgrey}(0.17)}$ ${\color{green}\uparrow}$     \\
FedNova~\cite{wang2020tackling}   & $58.44_{\color{coldgrey}(0.06)}$  ${\color{red}\downarrow}$  & $68.34_{\color{coldgrey}(1.67)}$ ${\color{green}\uparrow}$  & $68.65_{\color{coldgrey}(0.18)}$  ${\color{red}\downarrow}$ & $10.31_{\color{coldgrey}(76.05)}$ ${\color{red}\downarrow}$  & $96.60_{\color{coldgrey}(0.71)}$ ${\color{red}\downarrow}$  &  $98.96_{\color{coldgrey}(0.11)}$ ${\color{red}\downarrow}$   \\

FedAvgM~\cite{hsu2019measuring} & $51.49_{\color{coldgrey}(7.01)}$ ${\color{red}\downarrow}$  & $59.34_{\color{coldgrey}(7.33)}$ ${\color{red}\downarrow}$  & $56.60_{\color{coldgrey}(12.23)}$ ${\color{red}\downarrow}$
& $30.85_{\color{coldgrey}(55.51)}$ ${\color{red}\downarrow}$ & $97.51_{\color{coldgrey}(0.20)}$ ${\color{green}\uparrow}$ & $98.49_{\color{coldgrey}(0.58)}$ ${\color{red}\downarrow}$ \\

MOON~\cite{li2021model}   & $59.78_{\color{coldgrey}(1.72)}$ ${\color{red}\downarrow}$  & $98.49_{\color{coldgrey}(0.43)}$  ${\color{green}\uparrow}$ & $69.33_{\color{coldgrey}(0.50)}$  ${\color{green}\uparrow}$  & $77.71_{\color{coldgrey}(8.65)}$ ${\color{red}\downarrow}$ & $84.52_{\color{coldgrey}(12.79)}$  ${\color{red}\downarrow}$&$98.72_{\color{coldgrey}(0.35)}$ ${\color{red}\downarrow}$ \\

FedGKD~\cite{yao2021local}  & $58.08_{\color{coldgrey}(0.42)}$  ${\color{red}\downarrow}$ & $68.91_{\color{coldgrey}(2.24)}$ ${\color{green}\uparrow}$ & $69.00_{\color{coldgrey}(0.17)}$ ${\color{green}\uparrow}$ & $72.44_{\color{coldgrey}(13.92)}$  ${\color{red}\downarrow}$  & $88.06_{\color{coldgrey}(9.25)}$ ${\color{red}\downarrow}$ &  $99.23_{\color{coldgrey}(0.16)}$  ${\color{green}\uparrow}$  \\

FedProto~\cite{tan2022fedproto}  & $55.34_{\color{coldgrey}(3.16)}$  ${\color{red}\downarrow}$ & $70.04_{\color{coldgrey}(3.37)}$ ${\color{green}\uparrow}$ & $71.17_{\color{coldgrey}(2.34)}$ ${\color{green}\uparrow}$ & $32.02_{\color{coldgrey}(54.33)}$  ${\color{red}\downarrow}$  & $71.16_{\color{coldgrey}(27.61)}$ ${\color{red}\downarrow}$ &  $98.77_{\color{coldgrey}(0.30)}$  ${\color{red}\downarrow}$  \\

\midrule

\textbf{FedCSD (Ours)} & {$\textbf{60.15}_{\color{coldgrey}(1.65)}$} ${\color{green}\uparrow}$ & {$\textbf{71.36}_{\color{coldgrey}(4.69)}$} ${\color{green}\uparrow}$ & {$\textbf{71.53}_{\color{coldgrey}(4.86)}$}  ${\color{green}\uparrow}$ & {$\textbf{94.83}_{\color{coldgrey}(8.47)}$} ${\color{green}\uparrow}$  & {$\textbf{97.70}_{\color{coldgrey}(0.39)}$} ${\color{green}\uparrow}$  & {$\textbf{99.32}_{\color{coldgrey}(0.25)}$} ${\color{green}\uparrow}$ \\

\bottomrule
\end{tabular}}
}
\label{tab:table1}
\end{table*}

\subsection{Adaptive Mask}
We noted that the teacher is updated persistently during the whole training process, which is different from the normal knowledge distillation that uses a well-trained teacher to teach students on a dataset. Therefore, the distillation will slow the convergence and even make training collapse in the first few rounds due to the terrible soft labels of the teacher. Even though the performance of the teacher will be better as the training goes on, it still provides some wrong soft labels which are conflicting with the real labels, this lowers the upper bound of the method. To handle this problem, we filter out some terrible soft labels of teachers with an adaptive mask which can be defined as:
\begin{equation}
\small
\mathbb{M} \!=\! \begin{cases}
1,\!&\rho^t_{\xi,y_i}\!>\!\frac{1}{|\mathcal{Y}|} \\
0,\!&\text{otherwise}
\end{cases}, \ \text{where} \ \rho_{\xi,y_i}^{t}\!=\! \frac{\text{exp}(z_{\xi,y_i}^t)}{\sum_{i\in\mathcal{Y}}^{}\text{exp}(z_{\xi,y_i}^{t})},
\end{equation}
Notably, the mask is adaptively decided by the output class probability of the teacher. We argue that soft labels are worthless when the corresponding probability of the real class is smaller than $1/|\mathcal{Y}|$, this represents that the teacher does not yet have the ability to classify. 
With the proposed mask, the Eq.~\eqref{eq:lcsd} 
 can be rewritten as:
\begin{equation}
\mathcal{L}_{\mathrm{CSD}}=- \mathbb{E}_{(x_i,y_i)\sim \mathcal{D}_{k}}\mathbb{M}\tau^2\sum_{c\in\mathcal{Y}}^{}q_{\xi,c}^t\log (q_{k,c}^t),
\end{equation}

\subsection{Local Objective}
The class prototype similarity distillation loss is combined with the cross-entropy loss as the final local objective function, which can be described as:
\begin{equation}
\mathcal{L}^k = \mathcal{L}_{CE}^k + \mu\mathcal{L}_{CSD}^k,
\end{equation}
where $\mu$ is a hyper-parameter to control the contribution of the distillation loss.

\section{Experiments}

\subsection{Experimental Setup}

\noindent\textbf{{Datasets.}} We conduct extensive experiments on three typical datasets: \textbf{CIFAR-100}~\cite{krizhevsky2009learning}, \textbf{FEMNIST}~\cite{caldas2018leaf}, and \textbf{Office-Caltech-10}~\cite{gong2012geodesic}, which are widely used in FL~\cite{li2021model, yao2021local, zhou2023fedfa, tan2022fedproto}. To explore the generality of our method for non-IID data, we conduct experiments on two types of non-IID settings, \ie, label skew and feature skew~\cite{li2022federated}.
\begin{itemize}
	\setlength{\itemsep}{4pt}
	\setlength{\parsep}{-2pt}
	\setlength{\parskip}{-0pt}
	\setlength{\leftmargin}{-15pt}
	\item \textbf{Label Skew:} we adopt the Latent 
 Dirichlet Allocation~\cite{li2021model, yao2021local} strategy to divide the train set of CIFAR-100 and FEMNIST. Each client has an unbalanced number of categories under the above partitioning strategy. The data distribution $Dir(\beta)$ is controlled by the parameter $\beta$ and smaller $\beta$ has higher data heterogeneity. The $\beta$ is set as CIFAR-100 $\{0.01, 0.5, 5\}$ and FEMNIST $\{0.01, 0.05, 0.5\}$. The number of clients is set to 10 with the participation rate of 1 as default.
 	\item \textbf{Feature Skew:} following the previous work~\cite{zhou2023fedfa, li2021fedbn}, we adopt four different subsets of Office-Caltech 10: Amazon, Caltech, DSLR, and Webcam as 4 clients, which are from four different domains.
\end{itemize}
\noindent\textbf{{Implementation Details.}} We use ResNet-50~\cite{he2016deep} and Alexnet~\cite{krizhevsky2017imagenet} as classification networks for CIFAR100 and Office-Caltech-10, respectively. As for the easily classified FEMNIST, we use a simple convolutional neural network~\cite{li2021model} as the classification network. Our method and other baselines are implemented by PyTorch. Besides, we train all methods on a single NVIDIA GTX 1080Ti GPU with 11GB of memory. The batch size is 64 for CIFAR-100 and FEMNIST, and 32 for Office-Caltech-10. The SGD optimizer with a learning rate of 0.1 is used for all methods and the momentum and weight decay are set to 0.9 and 0.00001, respectively. The number of communication rounds is 100 with 5 local epochs each round for three datasets. For a fair comparison, we train all methods in the same environment and ensure that all methods have converged.

\noindent\textbf{{Baselines.}} We compare our proposed method with various state-of-the-art approaches include:
\begin{itemize}
	\setlength{\itemsep}{4pt}
	\setlength{\parsep}{-2pt}
	\setlength{\parskip}{-0pt}
	\setlength{\leftmargin}{-15pt}
	\item \textbf{FedAvg}~\cite{mcmahan2017communication}: a classical method in FL that averages directly all local model parameters.
	
	\item \textbf{FedProx}~\cite{li2020federated}: a method that improves FedAvg by introducing a proximal term into the local objective.
	
	\item \textbf{FedNova}~\cite{wang2020tackling}:  it normalizes and scales local updates at the weight average phase of FedAvg.

 	\item \textbf{FedAvgM}~\cite{hsu2019measuring}:  it introduces momentum to update the global model during the model aggregation.
	
	\item \textbf{MOON}~\cite{li2021model}: it pulls the representation of the current local model close to the global model and far away from the previous local model.
	
	\item \textbf{FedGKD}~\cite{yao2021local}: it integrates several global models of previous rounds as a teacher to regulate the local optimization during the local training. 
        \item \textbf{FedProto}~\cite{tan2022fedproto}: a prototype-based FL method, which aligns the global prototype and the latent feature of the local model during the local training. 
 
\end{itemize}
Notably, there are some key hyper-parameters in some baselines. For example, the loss function of FedProx, MOON, FedGKD, and FedProto is similar to us, which can be expressed as $\mathcal{L} = \mathcal{L}_{1} + \mu \mathcal{L}_{2}$. The $\mathcal{L}_{1}$ is the supervised loss term and $\mathcal{L}_{2}$ is an additional loss term proposed by their method. We fine-tune the $\mu$ from \{0.001, 0.01, 0.1, 1\} and report the best result for all methods. The optimal $\mu$ for FedProx, MOON, FedGKD, and FedProto is 0.001, 1, 0.01, and 1, respectively.  For other hyper-parameters, e.g., temperature parameter $\tau$, we adopt the best setting in their paper. In addition, for MOON, we discard the projection layer to keep the model consistent for a fair comparison. 
\begin{table}[t]
\centering
\caption{ \textbf{The test accuracy (\%) of all approaches on office-Caltech-10~\cite{gong2012geodesic}}. For a detailed comparison, we present the test accuracy of four clients: A(Amazon), C(Caltech), D(DSLR), W(Webcam), and the average result. ${\color{green}\uparrow}$ and ${\color{red}\downarrow}$ show the rise and fall compared with FedAvg. We mark the best results in bold.}
 \small
\setlength\tabcolsep{5pt}
\resizebox{0.5\textwidth}{!}{
\renewcommand\arraystretch{1.5}{
\begin{tabular}{llllll}
\toprule
\multirow{2}{*}{\textbf{Method}} & \multicolumn{5}{c}{\textbf{Office-Caltech-10}~\cite{gong2012geodesic}}  \\
\cmidrule(lr){2-6}
& A & C&D&W&Average\\
\midrule
FedAvg~\cite{mcmahan2017communication}   &$53.12$ & $44.88$ & $65.62$ & $86.44$ & $62.51_{\color{coldgrey}(base)}$ \\
\midrule
FedProx~\cite{li2020federated}    & $53.12$ & $\textbf{45.33}$ & $62.50$ & $86.44$ & $61.84_{\color{coldgrey}(0.67)}$ ${\color{red}\downarrow}$ \\
FedNova~\cite{wang2020tackling}   & $50.00$ & $42.22$ & $62.50$ & $\textbf{88.13}$ & $60.71_{\color{coldgrey}(1.80)}$ ${\color{red}\downarrow}$   \\

FedAvgM~\cite{hsu2019measuring} & $48.43$ & $\textbf{45.33}$ & $62.50$ & $83.05$ & $59.83_{\color{coldgrey}(2.68)}$ ${\color{red}\downarrow}$ \\

MOON~\cite{li2021model}   & $53.10$ & $44.88$& $\textbf{68.75}$ & $\textbf{88.13}$ & $63.20_{\color{coldgrey}(0.69)}$ ${\color{green}\uparrow}$  \\
FedGKD~\cite{yao2021local}  & $51.04$ & $44.00$ & $\textbf{68.75}$ & $84.74$ & $62.13_{\color{coldgrey}(0.38)}$ ${\color{red}\downarrow}$\\

FedProto~\cite{tan2022fedproto}  & $\textbf{55.72}$ & $44.44$ & $\textbf{68.75}$ & $86.44$ & $63.84_{\color{coldgrey}(1.33)}$ ${\color{green}\uparrow}$\\

\midrule

\textbf{FedCSD (Ours)} & $55.20$ & $\textbf{45.33}$ & $\textbf{68.75}$ & $\textbf{88.13}$ & $\textbf{64.35}_{\color{coldgrey}(1.84)}$ ${\color{green}\uparrow}$ \\

\bottomrule
\end{tabular}}
}
\label{tab:table2}
\end{table}

\noindent\textbf{{Detailed Setting of Our Method.}} The loss weight $\mu$ and temperature hyper-parameter are set to $0.001$ and $10$ for CIFAR-100 and FEMNIST. For Office-Caltech-10, the two hyper-parameters are set to $0.5$ and $4$, respectively. Besides, the momentum $\alpha$ is set to $0.9$ on three datasets by default.

\begin{table}[t]
\centering
\caption{\textbf{Ablation study of the key components} in our method on CIFAR-100~\cite{krizhevsky2009learning} and $\beta=0.5$, where $\hat{\delta}$ is class prototype similarity weighted score, $\mathbb{M}$ is adaptive mask.}
\setlength\tabcolsep{6pt}
\renewcommand\arraystretch{1.5}{
\small
\begin{tabular}{lcccc}
\toprule
\textbf{Method} & $\hat{\delta}$ & $\mathbb{M}$ & TMA & Accuracy\\
\midrule
FedAvg & - & - & - &  $66.67$\\
\midrule
Base & \XSolidBrush & \XSolidBrush & \XSolidBrush &  $63.66$\\
$\mathcal{M}_1$ & \Checkmark& \Checkmark &  \XSolidBrush & $70.19$ \\
$\mathcal{M}_2$  & \Checkmark& \XSolidBrush & \Checkmark & $69.38$\\
$\mathcal{M}_3$ & \XSolidBrush & \Checkmark & \Checkmark & $68.34$\\
\textbf{FedCSD (ours)}  & \Checkmark & \Checkmark & \Checkmark & $\textbf{71.36}$ \\
\bottomrule
\end{tabular}}
\label{tab:table3}
\end{table}
\begin{figure}[h]
\centering
  \includegraphics[width=0.48\textwidth]{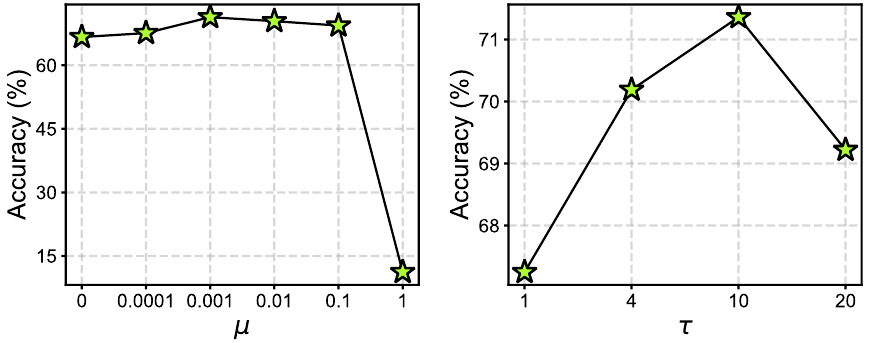}
  \caption{ \textbf{Illustration of test accuracy versus loss weight $\mu$ and $\tau$} on CIFAR-100~\cite{krizhevsky2009learning} and $\beta=0.5$.}
  \label{fig:para} 
\end{figure}
\begin{figure}[!t]
\centering
  \includegraphics[width=0.25\textwidth]{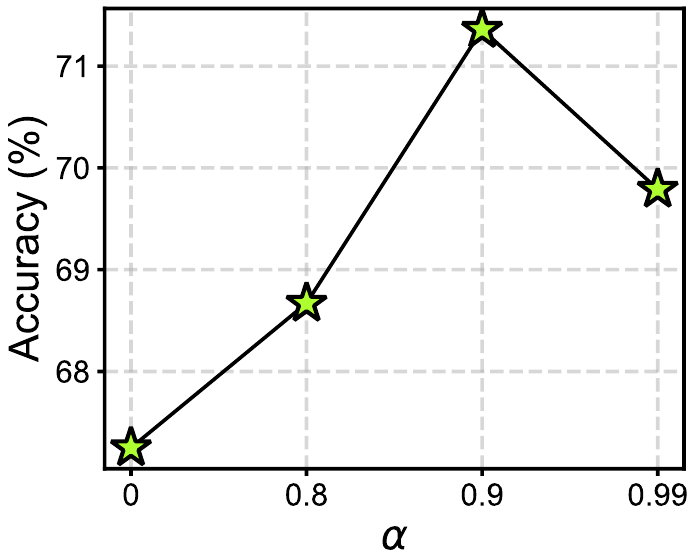}
  \caption{ \textbf{Analysis of the different values of $\alpha$} on CIFAR-100~\cite{krizhevsky2009learning} when $\beta=0.5$.}
  \label{fig:alpha} 
\end{figure}

\subsection{Accuracy Comparison}
We present the overall results on three benchmarks with two different non-IID settings. 

\noindent\textbf{{Results on Label Skew Setting.}} For this setting, we train all methods on the divided train set and evaluate them on the test set. Table~\ref{tab:table1} shows the experiment results of all methods on CIFAR-100 and FEMNIST with label skew non-IID data. As we can see, FedCSD achieves the best performance, which yields a consistent performance increase compared with other methods over different $\beta$. Particularly, FedCSD improves the accuracy of FedAvg as large as $4.69\%$ and $8.47\%$ on CIFAR-100 and FEMNIST, respectively. And compared with the methods that improve the local training phase, \ie, FedProx, and MOON, it also has significant improvements especially when data is highly heterogeneous. Besides, in comparison to the previous distillation-based method (FedGKD), it has a better performance, which confirms our method achieves more efficient distillation. FedCSD is also superior to the prototype-based method (Fedproto), which aligns the latent features of the local model and the prototype at the feature level, which shows the superiority of the logit solution. Notably, our method has a great improvement over other methods on FEMNIST when $\beta=0.01$, and the improvement is low due to the limited data heterogeneity when $\beta$ is higher.

\noindent\textbf{{Results on Feature Skew Setting.}} We present the accuracy of all methods on Office-Caltech-10 under the feature skew setting in Table~\ref{tab:table2}.  Different from the label skew setting, we evaluate all methods on the test set of four subsets. For a comprehensive comparison, we report the results of each client. Apparently, FedCSD achieves the best performance globally and locally. In contrast, other methods just can achieve the best performance locally. The results show that our method can address the feature skew non-IID data.

In general, judging from the above results, FedCSD is superior for non-IID data compared with other methods and has stronger generality for different non-IID settings.

\subsection{Ablation Studies}
\noindent\textbf{Influence of Key Components.} For a more detailed analysis of our methodology, we explore the influence of three components:  class prototype similarity weighted score $\hat{\delta}$, adaptive mask $\mathbb{M}$, and TMA. Therefore, we build up a new baseline, denoted \textbf{Base}, that directly distills the logits of the last round global model to the local model without our three components. And three baselines $\mathcal{M}_1$, $\mathcal{M}_2$, and $\mathcal{M}_3$ combine two of these components. The results of these methods on CIFAR-100 with $\beta = 0.5$ are presented in Table~\ref{tab:table3}. From the results, we can see that the accuracy of Base is even lower than the FedAvg due to the impact of the poorly trained global model and it can not achieve effective distillation. Besides, the accuracy of the three baselines, \ie, $\mathcal{M}_1$, $\mathcal{M}_2$, and $\mathcal{M}_3$, is declined to a certain degree compared with the full version of our method, which shows the importance of these three components. In particular, $\hat{\delta}$ has the most significant impact because it enhances the class similarity of teacher logits.

\begin{figure}[t]
\centering
  \includegraphics[width=0.48\textwidth]{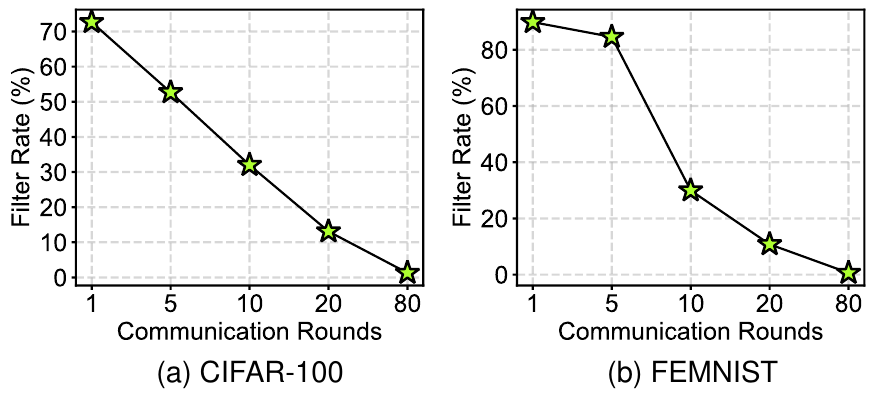}
  \caption{ \textbf{The filter rate of mask} versus communication rounds on CIFAR-100~\cite{krizhevsky2009learning} and FEMNIST~\cite{caldas2018leaf}.}
  \label{fig:mask} 
\end{figure}

\noindent\textbf{Influence of $\mu$ and $\tau$.} We explore the influence of two hyper-parameters: loss weight $\mu$ and temperature $\tau$ in our method. $\mu$ is tuned from \{$0.001, 0.01, 0.1, 1$\} and the range of $\tau$ is  \{$1, 4, 10, 20$\}. When we tune $\mu$ and $\tau$, the $\tau$, and $\mu$ are set to 10 and 0.001 as default, respectively. Thus, there is only a single variable in the experiments and the results are shown in Fig.~\ref{fig:para}. As shown in the figure, our method achieves the best accuracy when $\mu=0.001$ and $\tau=10$. Moreover, the accuracy is greatly dropped with large $\mu$ ($\mu=1$), which is attributed to that the distillation hinders the update of the local model. Yet it still can not obtain optimal accuracy when the contribution of the distillation is too small. As for $\tau$, the large value is better ($\tau=10$) because it can make the logits smoother, which is beneficial to distillation as the teacher can not provide reliable soft labels~\cite{hinton2015distilling}. However, it will weaken the knowledge of the teacher logits with too large $\tau$, which degrade the accuracy of the method.

\noindent\textbf{Influence of $\alpha$.} To explore the influence of $\alpha$ in our method, we tune $\alpha$ from $\{0, 0.8, 0.9, 0.99\}$ while $\mu$ and $\tau$ are set to $0.001$ and $10$ by default. As presented in Fig.~\ref{fig:alpha}, FedCSD yields the best result when $\alpha=0.9$. Besides, compared with $\alpha=0$, the accuracy of our method is increased when $\alpha > 0$. This indicates that TMA is beneficial to our method, which can provide a more stable teacher model.

\begin{figure*}[t]
\centering
  \includegraphics[width=0.96\textwidth]{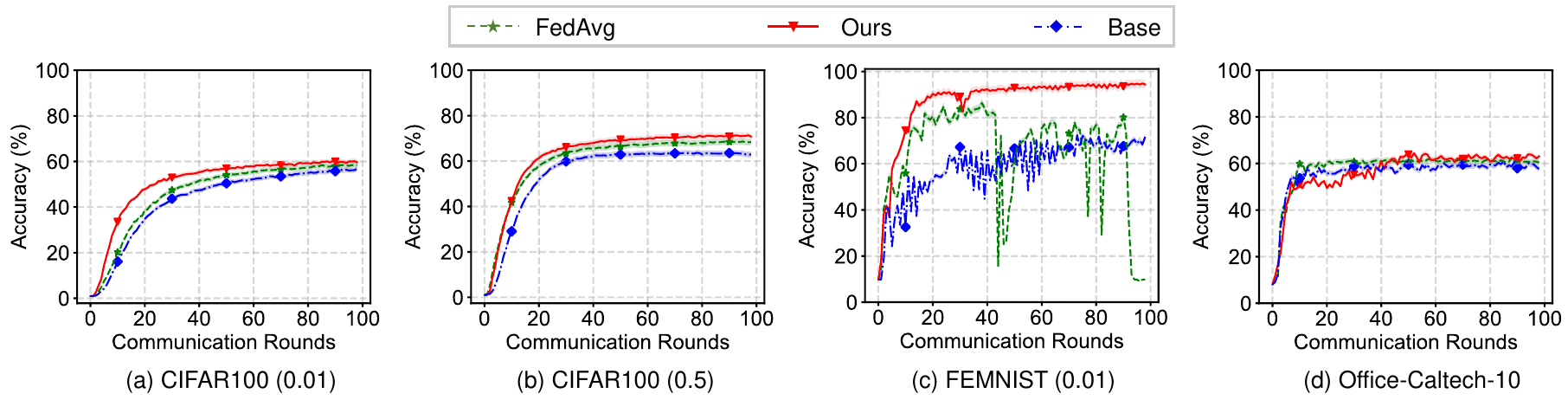}
  \caption{ \textbf{Illustration of test accuracy versus communication rounds} on CIFAR-100~\cite{krizhevsky2009learning}, FEMNIST~\cite{caldas2018leaf}, and Office-Caltech-10~\cite{gong2012geodesic}. }
  \label{fig:comm}
\end{figure*}
\begin{figure*}[t]
\centering
  \includegraphics[width=0.96\textwidth]{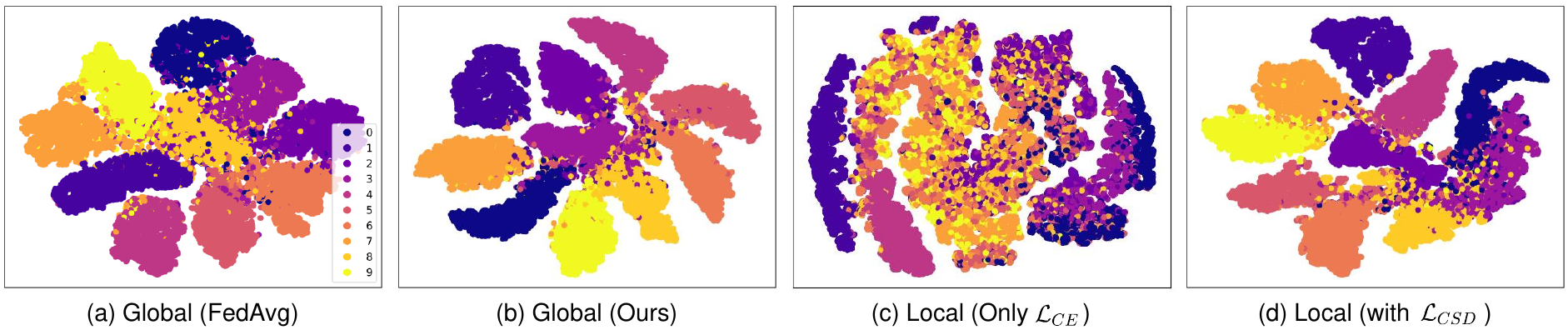}
  \caption{ \textbf{T-SNE~\cite{van2008visualizing} visualization of latent features} on FEMNIST~\cite{caldas2018leaf}. The global features of (a) FedAvg and  (b) Ours, and the local features of ours learned by (c) $\mathcal{L}_{CE}$ only, and (d) full version. }
  \label{fig:tsne}
\end{figure*}

\subsection{Analysis of Mask Filter.} 

To explore the underlying mechanism of the adaptive mask, we visualize its filter rate for wrong soft labels versus the communication round in Fig.~\ref{fig:mask}. As presented in the figure, the function of the mask is mainly in the early stage, which can effectively filter out the wrong soft labels. In the late, the mask can decrease its filter rate with the performance improvement of the global model, which preserves some valuable wrong soft labels. The mechanism of the mask fits the trait of the training process perfectly. We also compared the adaptive mask with another type of mask that filter out all wrong soft labels. 

\begin{table}[!t]
\centering
\caption{\textbf{ The test accuracy (\%) of different masks} on CIFAR-100~\cite{krizhevsky2009learning} with different $\beta$.}
 \small
\setlength\tabcolsep{8pt}
\resizebox{0.45\textwidth}{!}{
\renewcommand\arraystretch{1.5}{
\begin{tabular}{lccc}
\toprule
\multirow{2}{*}{\textbf{Method}} & \multicolumn{3}{c}{\textbf{CIFAR-100}~\cite{krizhevsky2009learning}}\\
\cmidrule(lr){2-4}
& $\beta=0.01$ & $\beta=0.5$ & $\beta=5$\\
\midrule

FedCSD + $\tilde{\mathbb{M}}$ & 59.95 & 70.03 & 69.61 \\
\textbf{FedCSD + $\mathbb{M}$ (Ours)} & \textbf{60.15} & \textbf{71.36} & \textbf{71.53}\\

\bottomrule
\end{tabular}}
}
\label{tab:diff_mask}
\vspace{-10pt}
\end{table}

\noindent\textbf{Different Types of Mask Filter.} To further explore the influence of our proposed adaptive mask, we compare it with another type of mask, which can be described as:
\begin{equation}
\tilde{\mathbb{M}} \!=\! \begin{cases}
1,\!& argmax(\rho^t_{\xi})\!= y_i \\
0,\!&\text{otherwise}
\end{cases}. \tag{16}
\end{equation}
Obviously, $\tilde{\mathbb{M}}$ is a forcible mask that removes all the wrong soft labels, which will filter out the valuable knowledge. We present the comparative result in Table.~\ref{tab:diff_mask}. As we can see, the adaptive mask $\mathbb{M}$ is superior to the forcible mask $\tilde{\mathbb{M}}$ in various settings. Notably, $\tilde{\mathbb{M}}$ achieves similar performance with $\mathbb{M}$ due to the low accuracy of the global model, indicating the forcible mask is a feasible strategy in this case that can filter out more wrong knowledge. 
However, compared with $\beta=0.5$, the forcible mask $\tilde{\mathbb{M}}$ even achieves lower accuracy under $\beta=5$. Because the accuracy of the global model is higher and improves the quality of soft labels, $\tilde{\mathbb{M}}$ will filter out the valuable knowledge of wrong soft labels. 

\subsection{Communication Efficiency}
\noindent\textbf{Convergence Rate.} To explore the convergence rate of our method, we draw the test accuracy curve with different communication rounds as shown in Fig.~\ref{fig:comm}. Apparently, our method has a better convergence compared FedAvg in the label skew setting (Fig.~\ref{fig:comm} (a, b, c)). Especially in (Fig.~\ref{fig:comm} (c)), thanks to the elaborate design, our method has a more stable convergence process compared with Base and FedAvg. In the feature skew setting (Fig.~\ref{fig:comm} (d)), the convergence of FedCSD is slightly lower than FedAvg, yet it improves the upper bound of the accuracy.

\noindent\textbf{Communication Cost.} We note that the acquisition of the teacher (Eq.~\ref{eq:tma}) can be put into each client. The client can use some memory to store the teacher model and update it with the received global model before the local training. Therefore, it only increases the communication cost of the prototype compared with FedAvg. However, the cost of the prototype is tiny because it is a $|\mathcal{Y}| \times |\mathcal{Y}|$ matrix, where $|\mathcal{Y}|$ is the number of class, \eg, 10 for FEMNIST.

\subsection{Feature Distribution}
We show the learned features of the global model trained by FedAvg and our method in Fig~\ref{fig:tsne} (a, b).  Compared with FedAvg, our method has a better feature representation in that the features from the same class are clustered and separated well, thus the classifier can easy to learn the decision boundary to identify them. Besides, as stated in \S\ref{sec:moti}, our method can mitigate the feature logits by keeping the logits consistent, which is observed from the exploratory experiment. Therefore, we conduct an additional experiment that uses the global model parameters (Fig~\ref{fig:tsne} (b)) to initialize the local model and the local model is then trained on the local dataset in two ways. One is trained with the cross-entropy loss $\mathcal{L}_{CE}$ only and another is trained with our local loss function. We visualize the features of two local models in Fig~\ref{fig:tsne} (c, d). As we can see,  the features of the local model learned from $\mathcal{L}_{CE}$ are mixed, which shows that it has lost the ability to classify some classes due to bias in the skew local dataset. In contrast, our local model still has good classification boundaries, which indicates the local model remains the global knowledge instead of biasing in the local dataset during the local training. In a nutshell, the above results confirmed that our method can mitigate the feature difference between the local and global models.

\section{Conclusion}
In this work, we focused on addressing the client drift problem caused by non-IID data. We observed that the difference between local and global logits is positively correlated with the local epochs, which decreases the accuracy of FedAvg. Motivated by this, we proposed a new FL method, FedCSD, which explored a new perspective, the relation of local-global logits, to mitigate client drift. Our experiments show that FedCSD achieves significant improvement over FedAvg and outperforms other state-of-the-art methods in different data settings.

{\small
\bibliographystyle{IEEEtran}
\bibliography{tnnls_fedcsd}
}

\end{document}